\documentclass[a4paper, 10 pt, conference]{ieeeconf}
\IEEEoverridecommandlockouts

\usepackage[usenames,dvipsnames]{color}
\usepackage{graphicx}
\usepackage{cite}
\usepackage{tikz}
\usepackage{amsmath}
\usepackage{mathtools}
\usetikzlibrary{patterns}
\usepackage{bbm}
\usepackage{bm}
\usepackage{amsfonts}

\graphicspath{{images/}}

\usepackage{xargs}
\usepackage{verbatim}
\usepackage{graphics}
\usepackage{epsfig} 
\usepackage{subcaption}

\usepackage{xargs}
\usepackage{verbatim}
\usepackage{graphics}
\usepackage{epsfig} 

\usepackage{amsmath,lipsum} 
\usepackage{amssymb}  
\usepackage{xfrac}

\definecolor{wheat}{rgb}{0.96,0.87,0.70}
\definecolor{tommy}{rgb}{0.8,0.8,1}
\definecolor{ivo}{rgb}{1,0.8,0.8}

\title{\LARGE \bf
	Learning When to Drive in Intersections by Combining Reinforcement Learning and Model Predictive Control
}

\author{Tommy Tram$^{1,2,3}$, Ivo Batkovic$^{1,2,3}$, Mohammad Ali$^{1}$, and Jonas Sj\"oberg$^{2}$
	\thanks{This work was partially supported by the Wallenberg AI, Autonomous Systems and Software Program (WASP) funded by the Knut and Alice Wallenberg Foundation.}
	\thanks{$^{1}$ Affiliated with Zenuity AB, Gothenberg, Sweden
		{\tt\small \{tommy.tram, ivo.batkovic, mohammad.ali\}@zenuity.com}}%
	\thanks{$^{2}$ Affiliated with the Department of Electrical Engineering, Chalmers University of Technology, Gothenberg, Sweden
		{\tt\small \{tommy.tram, ivo.batkovic, jonas.sjoberg\}@chalmers.se}}%
	\thanks{$^{3}$ Affiliated with AI Innovation of Sweden, Gothenburg, Sweden}%
}

\newcommand {\matr}[2]{\left[\begin{array}{#1}#2\end{array}\right]}

\newcommand{\x}{{\mathbf{x}}}
\renewcommand{\u}{{\mathbf{u}}}

\renewcommand{\r}{{\mathbf{r}}}
\begin{document}

\maketitle

\begin{abstract}
	In this paper, we propose a decision making algorithm intended for automated vehicles that negotiate with other possibly non-automated vehicles in intersections. The decision algorithm is separated into two parts: a high-level decision module based on reinforcement learning, and a low-level planning module based on model predictive control. Traffic is simulated with numerous predefined driver behaviors and intentions, and the performance of the proposed decision algorithm was evaluated against another controller. The results show that the proposed decision algorithm yields shorter training episodes and an increased performance in success rate compared to the other controller. 
\end{abstract}


%
\IEEEpeerreviewmaketitle

\section{Introduction}
\noindent
How can a self-driving vehicle interact and drive safely through intersections with other road users?
Interactions between road users in intersections is a complex problem to solve, making it difficult to address using conventional rule based systems.
Many advancements aim to solve this problem by trying to imitate human drivers \cite{Bansal2018ChauffeurNet:Worst} or predicting what other drivers in traffic are planning to do \cite{Zyner2017LongPrediction}. In \cite{BrechtelProbabilisticPOMDPs}, the authors show that by modeling the decision process as a partially observable Markov decision process, the model can account for uncertainty in sensing the environment and \cite{BoutonReinforcementDriving} showed some probabilistic guarantees when solving the problem using reinforcement learning (RL). 

\begin{figure}[t!]
	\centering
	\vspace{0.3cm}
	\includegraphics[width=0.8\columnwidth]{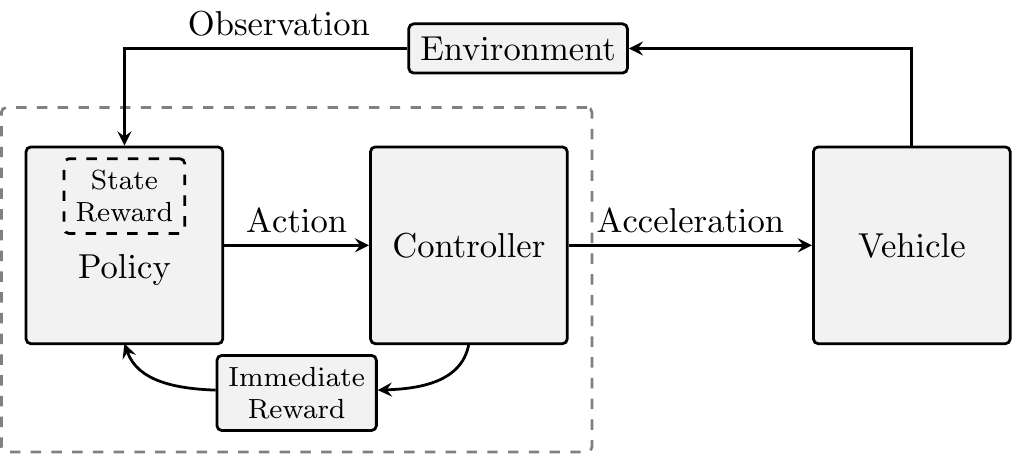}
	\caption{Representation of the decision making architecture. The dashed line marks the decision algorithm that is separated into two parts; the high-level decision maker, denoted as the policy, and the low-level controller.}
	\label{fig:Architecture}
	\vspace{-0.3cm}
\end{figure}

Previous research \cite{Tram2018LearningQ-Learning} showed that reinforcement learning can be used to learn a negotiation behavior between vehicles without vehicle to vehicle communication when driving in an intersection. The method found a policy that could avoid collisions in an intersection with crossing traffic, where other vehicles have different intentions. 
Since the previous work separates the framework in a high-level decision maker and a low-level controller, the high-level decision making algorithm can focus on the task when to drive, while the low level controller handles the comfort of passengers in the car by generating a smooth acceleration profile. We showed how this worked for intersections with a single crossing point, where Short Term Goal (STG) actions could choose one car to follow. This architecture, similar to Fig.~\ref{fig:Architecture}, gives the decision algorithm, the RL policy, the flexibility to choose actions that can safely drive through the intersection by switching between numerous STG. 
A solution with a simple controller holds well when the distance between intersection are far away from each other, but when there are several crossing points in close succession, the system would have a hard time avoiding collisions due to the increased complexity of multiple points and timeing where a collision can occur. 

In this paper, we instead propose to combine the high-level decision maker from \cite{Tram2018LearningQ-Learning} with a Model Predictive Controller~(MPC) in a framework presented in Fig.~\ref{fig:Architecture}. The performance of the MPC controller is benchmarked against a Sliding Mode controller that was used in \cite{Tram2018LearningQ-Learning}. The benefit of the MPC is that it can consider multiple vehicles at the same time and generate an optimal trajectory, which instantly gives feedback on performance and feasibility, i.e. predicting collisions, to the high-level decision maker. In contrast to \cite{hult}, where the authors prove stability and recursive feasibility using an MPC approach and assuming that agents can cooperate, we restrict ourselves to non-cooperative scenarios. 

Applying MPC directly to the problem could lead to a growing complexity with the number of vehicles in the intersection, e.g., the vehicle needs to decide based on multiple options which vehicle to yield for, and which to drive in front of.  Therefore, we propose to separate the problem into two parts: the first being a high-level decision maker, which structures the problem, and the second being a low level planner, which optimizes a trajectory given the traffic configuration. 

For the high-level decision maker, RL is used to find a policy for how the vehicle should drive through the intersection, and MPC is used as a low-level planner to optimize a safe trajectory. Compared to \cite{decentralizedMPC,de2013autonomous} where all vehicles are controlled using MPC to stay in safe sets, based on models of other vehicles' future trajectory, this could possibly be perceived as too conservative for a passenger. By combining RL and MPC, the decision policy will learn which action is optimal by using feedback from the MPC controller in the reward function. Since MPC uses predefined models, e.g. vehicle models and other obstacle prediction models, the performance relies on their accuracy and assumptions. To mitigate this, we use Q-learning, a model-free RL approach, to maximize the expected future reward based on the experience gained during an entire episode. This approach is able to compensate, to some extent, for model errors and is explained more in Section \ref{sec:q-learning}. 

In this work, we focus on the integration between policy and actuation, by having an MPC controller directly giving feedback to the decision maker through immediate reward, allowing the policy to know how comfortably the controller can handle an action and give feedback sooner if the predicted outcome may be good or bad. 


This paper is structured as follows. Section \ref{sec:problem} introduces the problem formulation along with the two-layers of the decision algorithm. Section \ref{sec:agents} presents three agents used for simulation and validation. Implementation details are presented in Section \ref{sec:implementation}, and the results are shown in Section \ref{sec:results} followed by discussion in Section \ref{sec:discussion}. Finally, we draw conclusions in Section \ref{sec:conclusion}.

\

\section{Problem formulation}\label{sec:problem}
\noindent
The goal of the ego-vehicle is to drive along a predefined route that has one or two intersections with crossing traffic, where the intent of other road users is unknown. Therefore, the ego-vehicle needs to assess the driving situation and drive comfortably, while avoiding collisions with any vehicle\footnote{Although our approach can be extended to other road users, for convenience of exposition we'll refer to vehicles.} that may cross. In this section, we define the underlying Partially Observable Markov Decision Process~(POMDP) and present how the problem is decomposed using RL for decision making and MPC for planning and control.


\subsection{Partially Observable Markov Decision Process}\label{sec:pomdp}
\noindent
A POMDP \cite{Kochenderfer2015} is defined by the 7-tuple $(\mathcal{S},\mathcal{A},\mathcal{T},\mathcal{R},\Omega ,\mathcal{O},\gamma)$, where $\mathcal{S}$ is the state space, $\mathcal{A}$ an action space that is defined in section \ref{sec:mpc}, $\mathcal{T}$ the transition function, 
the reward function $\mathcal{R}: \mathcal{S} \times \mathcal{A} \to \mathbb{R}$ is defined in \ref{sec:reward}, $\Omega$ an observation space, $\mathcal{O}$ the probability of being in state $s_{t}$ given the observation $o_t$, and $\gamma$ the discount factor.

A POMDP is a generalization of the Markov Decision Process (MDP) \cite{BellmanMDP} and therefore works in the same way in most aspects. At each time instant $t$, an action, $a_t\in \mathcal{A}$, is taken, which will change the environment state $s_t$ to a new state $s_{t+1}$. Each transition to a state $s_t$ with an action $a_t$ has a reward $r_t$ given by a reward function $\mathcal{R} $. The key difference from a regular MDP is that the environment state $s_t$ is not entirely observable, e.g., the intention of other vehicles is not known.
In order to find the optimal solution for our problem, we need to know the future intention of other drivers. Instead, we can only partially perceive the state though observations $o_t\in \Omega$.

\subsection{Q-Learning}\label{sec:q-learning_top}
\noindent
In the reinforcement learning problem, an agent observes the state $s_t$ of the environment, takes an action $a_t$, and receives a reward $r_t$ at every time step $t$. Through experience, the agent learns a policy $\pi$ in a way that maximizes the accumulated reward $\mathcal{R}$ in order to find the optimal policy $\pi^*$. In Q-learning, the policy is represented by a state action value function $Q(s_t,a_t)$. The optimal policy is given by the action that gives the highest Q-value. 
\begin{equation}
\pi^*(s_t) = \arg\max_{a_t} Q^*(s_t,a_t)
\label{eq:optimal_policy}
\end{equation}
Following the Bellman equation, the optimal Q-function $Q^*(s_t,a_t)$ is given by
 \begin{equation}
 Q^*(s_t,a_t)= \mathbb{E}[r_t + \gamma \max_{a_{t+1}} Q^*(s_{t+1}, a_{t+1})| s_t, a_t].
 \label{eq:q-function}
 \end{equation}
 
 \

\section{Agents}\label{sec:agents}
\noindent
The action space $\mathcal{A}$ is made out of six actions. The first two actions: $\alpha_1$ take way, and $\alpha_2$ give way, have the simple goal of crossing the intersection and stopping before the intersection, respectively. The actions $\alpha_{2+j}$ has the goal of following a vehicle $j$. 

In the following, we explain the two agents used for control of the ego vehicle and how they apply each action and how the surrounding traffic is modeled with varying intentions.

\subsection{MPC agent}\label{sec:mpc}
\begin{figure}[t]
	\centering
	\begin{subfigure}[t]{0.45\columnwidth}
		\centering
		\includegraphics[width=\columnwidth]{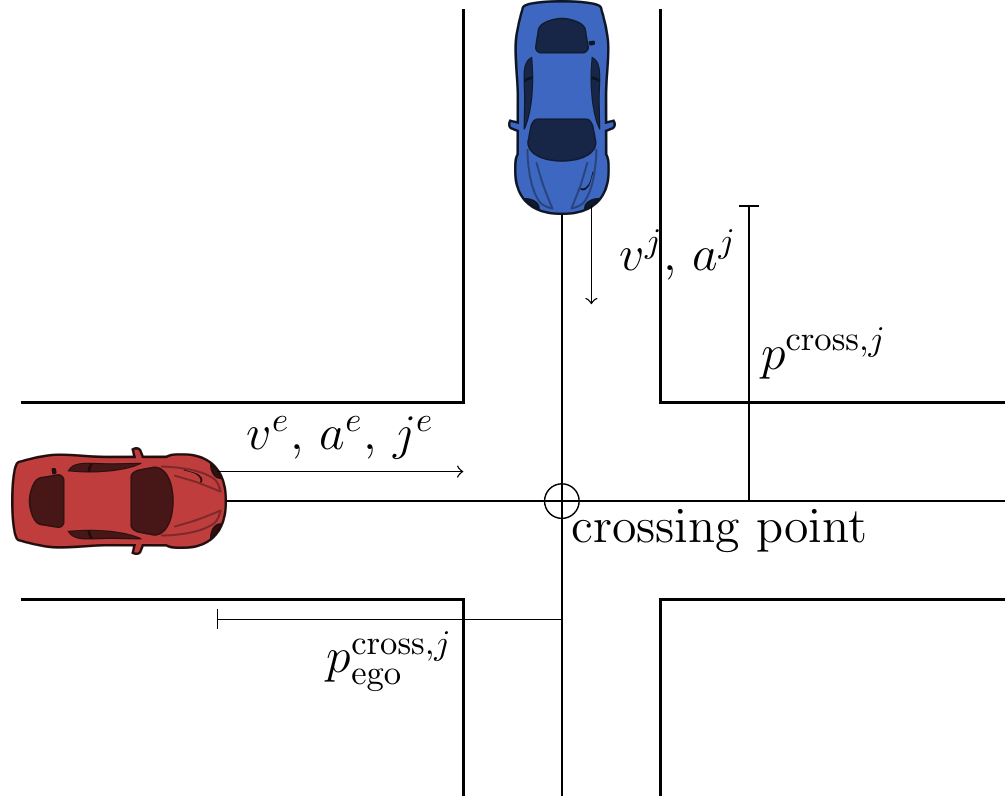}
		\caption{Observations}
		\label{fig:observations}
	\end{subfigure}
	\hfill
	\begin{subfigure}[t]{0.49\columnwidth}
		\centering
		\includegraphics[width=\columnwidth]{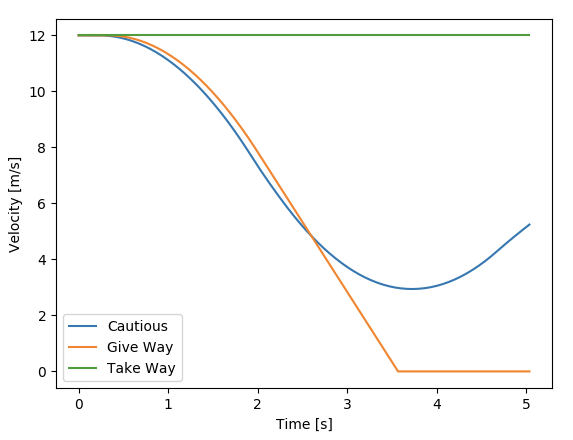}
		\caption{Velocity profile examples of various intention agents.}
		\label{fig:intention_profiles}
	\end{subfigure}
	\caption{Observations and agents in a scenario}
	\vspace{-0.5cm}
\end{figure}

We model the vehicle motion with states $\mathbf{x}\in\mathbb{R}^3$ and control $\mathbf{u}\in\mathbb{R}$, defined as
\begin{equation}
\mathbf{x}:=[p^\mathrm{e}\quad v^\mathrm{e}\quad a^\mathrm{e}]^\top,\quad \mathbf{u}:=j^\mathrm{e},
\end{equation}
where we denote the position along the driving path in a Frenet frame as $p^\mathrm{e}$, the velocity as $v^\mathrm{e}$, the acceleration as $a^\mathrm{e}$, and the jerk as $j^\mathrm{e}$, see Fig. \ref{fig:observations}. In addition, we assume  that measurements of other vehicles are provided through an observation  $\mathbf{o}$. We limit the scope of the problem to consider at most four vehicles, and define  the observations as
\begin{equation}
\mathbf{o} := [p^1\quad v^1\quad p^{\mathrm{cross},1}_\mathrm{ego}\quad \cdots\quad p^4\quad v^4\quad p^{\mathrm{cross},4}_\mathrm{ego}\quad ]^\top,
\end{equation}
where we denote the position along its path as $p^j$, the velocity as $v^j$, and $p^{\mathrm{cross},j}_\mathrm{ego}$ for $j\in[1,4]$, as the distance to the ego-vehicle from the intersection point, see Fig. \ref{fig:observations}. Note that we distinguish between $p^{\mathrm{cross},j}_\mathrm{ego}$, since vehicles can cross at different points, see Fig. \ref{fig:scenario}.

We assume that there exists a lateral controller that stabilizes the vehicle along the driving path. To that end, we only focus on the longitudinal control. Given the state representation, the dynamics of the vehicle is then modeled using a  triple integrator with jerk as control input.

The objective of the agent is to safely track a reference, i.e. follow a path with a target speed, acceleration, and jerk profile, while driving comfortably and satisfying constraints that arise from physical limitations and other road users, e.g. not colliding in intersections with crossing vehicles. Hence, we formulate  the  problem as a finite horizon, constrained optimal control problem
\begin{subequations}
	\label{eq:mpc}
	\begin{align}
	\min_{\bar\x,\bar\u} & \sum_{k=0}^{N-1}
	\matr{c}{\bar\x_k - \r_k^\x \\ \bar\u_k - \r_k^\u}^\top \matr{cc}{\bar{Q} &\bar{S}^\top\\\bar{S} & \bar{R}} \matr{c}{\bar\x_k - \r_k^\x \\ \bar\u_k - \r_k^\u} \\
	&\qquad + \matr{c}{\bar\x_N - \r_N^\x}^\top \bar{P} \matr{c}{\bar\x_N - \r_N^\x}\nonumber\\
	&\text{s.t.}\ \ \, \bar\x_0 = \hat{\x}_0, \label{eq:mpcState}\\
	&\qquad{}\bar\x_{k+1} = A\bar\x_{k}+B\bar\u_{k},\label{eq:mpcDynamics}\\
	&\qquad{}h(\bar\x_k,\bar\u_k,\bar{\mathbf{o}}_k,a) \leq{} 0, \label{eq:mpcInequality}
	\end{align}
\end{subequations}
where $k$ is  the prediction time index, $N$ is the prediction horizon, $\bar{Q}$, $\bar{R}$, and $\bar{S}$ are the stage costs, $\bar{P}$ is the terminal cost, $\bar\x_k$ and $ \bar\u_k$ are the predicted state and control inputs, $\r^\x_k$ and $\r^\u_k$ are the state and control input references, $\bar{\mathbf{o}}_k$ denotes the predicted state of vehicles in the environment which need to be avoided, and $a$ is the action from the high-level decision maker in Sec.~\ref{sec:q-learning_top}. Constraint \eqref{eq:mpcState} enforces that the prediction starts  at the current state estimate $\hat\x_0$, \eqref{eq:mpcDynamics} enforces the system dynamics, and \eqref{eq:mpcInequality} enforces constraints on the states, control inputs, and obstacle avoidance.

The reference points, $\r^\x_k$, $\r^\u_k$ are assumed to be set-points of a constant velocity trajectory, e.g. following the legal speed-limit of the road. Therefore, we set the velocity reference according to the speed limit, and the acceleration and jerk to zero.

\subsubsection{Obstacle prediction}
In order for the vehicle planner in \eqref{eq:mpc} to be able to properly avoid collisions, it is necessary to provide information about the surrounding vehicles in the environment. Therefore, similarly to \cite{batkovic2019}, we assume that a sensor system provides information about the environment, and that there exists a prediction layer which generates future motions of other vehicles in the environment. The accuracy of the prediction layer does indeed affect the performance of the planner, however, since the high-level decision maker is separated from the low level control, the decisions can still be made robust to handle model errors and prediction errors.

In this paper, for simplicity the future motion of other agents is estimated by a constant velocity prediction model. The motion is predicted at every time instant for prediction times $k\in[0,N]$, and is used to form the collision avoidance constraints, which we describe in the next section. Even though more accurate prediction methods do exist, e.g. \cite{lefevre2014survey,batkovic2018}, we use this simple model to show the potential of the overall framework.
\
\subsubsection{Collision avoidance}

We denote a vehicle $j$ with the following notation $\x^j:=[p^j\, v^j\, a^j]^\top$, and an associated crossing point at position $p^{\mathrm{cross},j}$ in its own vehicle frame, which translated into the ego-vehicle frame is denoted as $p^{\mathrm{cross},j}_\mathrm{ego}$. For clarity, see Fig. \ref{fig:observations}. With a predefined road topology, we assume that the vehicles will travel along the assigned paths, and that collisions may only occur at the intersection points $p^{\mathrm{cross},j}$ between an obstacle and the ego vehicle. Hence, for collision avoidance, we use the predictions of the future obstacle states $\bar\x^j_k$ for times $k\in[0,N]$, provided by the prediction layer outside of the MPC framework. Given the obstacle measurements, the prediction layer will generate future states throughout the prediction horizon. With this information, it is possible to identify the time slots when a vehicle enters and exits the intersection.

Whenever an obstacle $j$ is predicted to be within a threshold of $p^{\mathrm{cross},j}$, e.g. the width of the intersecting area, the ego vehicle faces a constraint of the following form
\begin{gather*}
\bar{p}_k^\mathrm{e} \geq{} p^{\mathrm{cross},j}_\mathrm{ego} + \Delta,\quad\underline{p}_k^\mathrm{e} \leq{} p^{\mathrm{cross},j}_\mathrm{ego} - \Delta,
\end{gather*}
where $\Delta$ ensures sufficient padding from the crossing point that does not cause a collision. The choice of $\Delta$ must be at least such that $p_k$ together with the dimensions of the ego-vehicle does not overlap with the intersecting area.

\subsubsection{Take way and give way constraint}
Since the constraints from the surrounding obstacles become non-convex, we rely on the high-level policy maker to decide through action $a\in\mathcal{A}$ how to construct constraint \eqref{eq:mpcInequality} for Problem \eqref{eq:mpc}. The take-way action implies that the ego-vehicle drives first through the intersection, i.e., it needs to pass the intersection before all other vehicles. This implies that for any vehicle $j$ that reaches the intersection during prediction times $k\in[0,N]$, the generated constraint needs to lower bound the state $p_k$ according to
\begin{equation}
\max_{j}p^{\mathrm{cross},j}+\Delta \leq{}p_k^\mathrm{e}.
\end{equation}
Similarly, if the action is to give way, then the position needs to be upper bounded by the closest intersection point so that
\begin{equation}
p_k^\mathrm{e} \leq{} \min_{j}p^{\mathrm{cross},j}_\mathrm{ego}-\Delta,
\end{equation} 
for all times $k$ that the vehicle is predicted to be in the intersection.

\subsubsection{Following an obstacle} If action $a\in\mathcal{A}$ is not chosen to give way, or to take way, the remaining options are to follow one of the $j$ vehicles. For such choices on $a$ the ego-vehicle position is upper bounded by $p^\mathrm{e}_k \leq{} p^\mathrm{cross,j}_\mathrm{ego}$. For other vehicles $i\neq{}j$, we construct the following constraints
\begin{itemize}
	\item if $p^\mathrm{cross,i}<{}p^\mathrm{cross,j}$ then $p^\mathrm{cross,i}+\Delta\leq{}p_k^\mathrm{e}$, which implies that the ego-vehicle should drive ahead of all vehicles $i$ that are approaching the intersection;
	\item if $p^\mathrm{cross,i}>{}p^\mathrm{cross,j}$ then $p_k^\mathrm{e}\leq{}p^\mathrm{cross,i}-\Delta$, which implies that the ego-vehicle should wait to pass vehicle $j$ and other vehicles $i$;
	\item if $p^{\mathrm{cross,i}}=p^{\mathrm{cross},j}$ then the constraints generated for vehicle $i$ becomes an upper or lower  bound depending on if vehicle $i$ is ahead or behind vehicle $j$ into the intersection.
\end{itemize}

\subsection{Sliding mode agent}
\noindent
To benchmark the performance of using MPC, we introduce a Sliding Mode~(SM) controller that was used in \cite{Tram2018LearningQ-Learning} and given by the following
\begin{subequations}
	\label{eq:sliding_mode}
	\begin{align}
	a^e_{sm} &= \frac{1}{c_2} (- c_1 x_2 + \mu sign(\sigma(x_1, x_2))), \\
	&\text{where}
	\begin{cases}
	x_1 = p^t - p^e,\\
	x_2 = v^t - v^e,
	\end{cases}\\
	&\sigma = c_1 x_1 + c_2 x_2,\\
	\label{eq:p-controller}
	&a^e_\mathrm{p} = K (v_{\mathrm{max}} - v^e),\\
	\label{eq:final_acc}
	&a^e = \min(a^e_{\mathrm{sm}}, a^e_\mathrm{p} ).
	\end{align}
\end{subequations}
The SM controller aims to keep a minimum distance to a target vehicle with a velocity of $v^e$, by controlling the acceleration $a^e_{\mathrm{sm}}$. The tuning parameters $c_1$, $c_2$, and  $\mu$  are used to tune the comfort of the controller. In case no target vehicle exists, the controller maintains a target velocity $v_{\mathrm{max}}$ with a proportional control law from \eqref{eq:p-controller} with the proportional constant $K$. The final acceleration is given by \eqref{eq:final_acc}. For more details about the SM agent see \cite{Tram2018LearningQ-Learning}.


\subsection{Surrounding traffic agents}
\noindent
There are three intentions for agents in surrounding traffic. Examples of some velocity profiles are shown in Fig. \ref{fig:intention_profiles}. The intention of all agents is implemented with a SM controller with various target values. The take way intention does not yield for the crossing traffic and simply aim to keep its target reference speed. The give way intention, however, slows down to a complete stop at the start of the intersection, until crossing traffic has passed. The third intention is cautious, i.e. slowing down but not to a full stop. 

\section{Implementation}\label{sec:implementation}

\subsection{Deep Q-Network}

\begin{figure}[t]
	\centering
	\includegraphics[width=0.95\columnwidth]{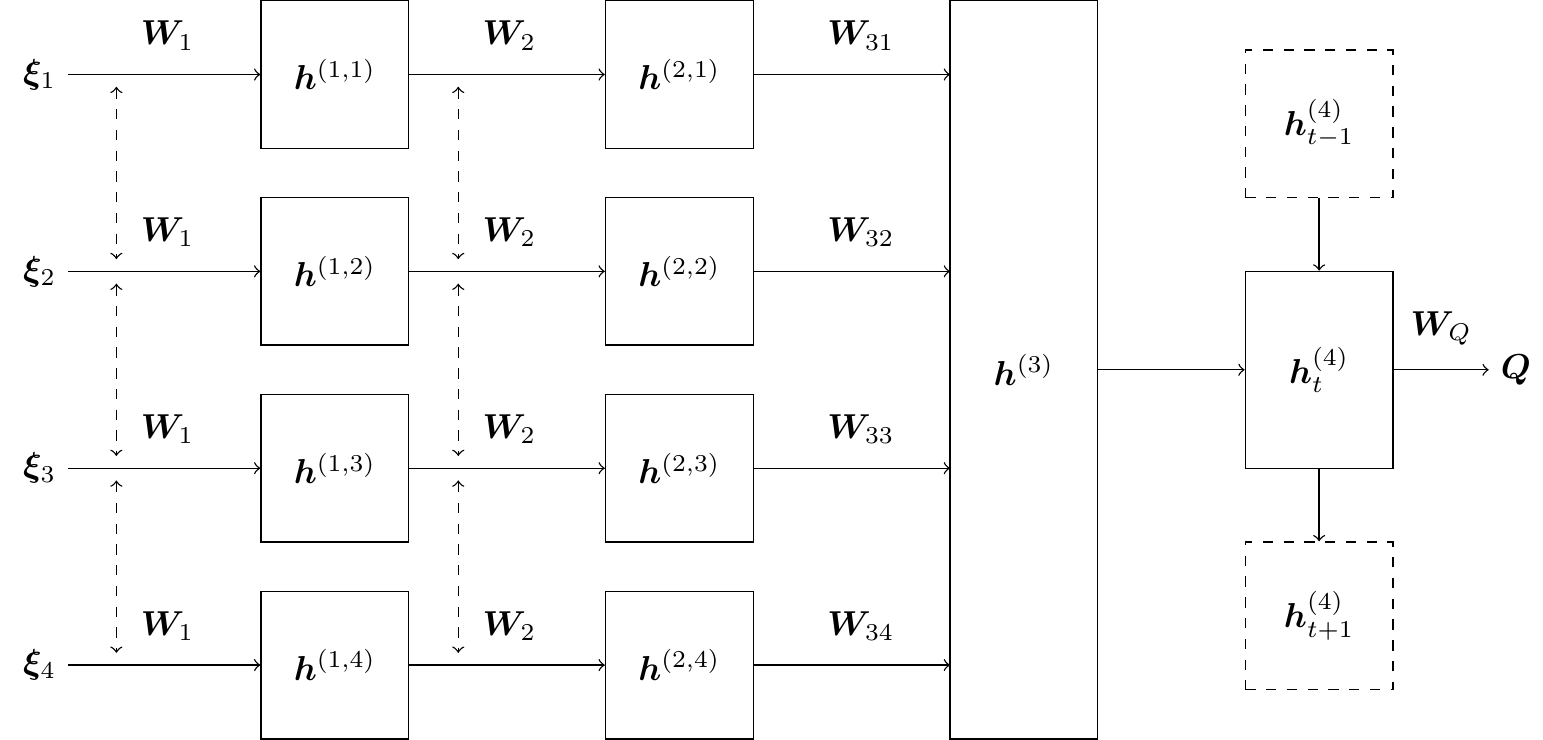}
	\caption{Representation of the network structure}
	\label{fig:Network}
\end{figure}
\label{sec:q-learning}
\noindent
The deep Q-network is structured as a three layer neural network with shared weights and a Long Short-Term Memory~(LSTM) layer based on previous work \cite{Tram2018LearningQ-Learning} and shown in Fig. \ref{fig:Network}. A similar study for lane changes on a highway confirmed the importance of having equal weights for inputs that describe the state of interchangeable objects \cite{Hoel}. The input features $\xi_n$ are composed of observations $o_t$, introduced in section \ref{sec:pomdp} and shown in Fig. \ref{fig:observations}, with up to four observed vehicles

\begin{equation}
	\xi_n = [\  p^e_t \quad v^e_t \quad a^e_t \quad \delta^e \quad p^n_t \quad v^n_t \quad a^n_t \quad \delta^n \  ]^T\\
\end{equation}

Normalization of the input features is done by scaling the features down to values between $[-1,1]$ using the maximum speed $v_{\max}$, maximum acceleration $a_{\max}$ and a car's sight range $p_{\max}$. Empty observations of other vehicles $[p^n_t \quad v^n_t \quad a^n_t \quad \delta^n ]$ has a default value of $-\mathbf{1}$. The input vectors are sent though two hidden layers $\bm{h}^{(1, i)}$ and $\bm{h}^{(2, i)}$ with shared weights $\bm{W}_1$ and $\bm{W}_2$ respectively
\begin{equation}
\bm{h}^{(1, i)} = \tanh\left(\bm{W}_1 \bm\xi_i + \bm{b}_1\right),
\end{equation}
\begin{equation}
\bm{h}^{(2, i)} = \tanh\left(\bm{W}_2 \bm{h}^{(1, i)} + \bm{b}_2\right).
\end{equation}
The output of each sub-network is then sent through a fully connected layer 
\begin{equation}
\label{eq:shared_weights}
\bm{h}^{(3)} = \tanh\left(\sum_{i=1}^4 \bm{W}_{3i} \, \bm{h}^{(2, i)} + \bm{b}_3\right),
\end{equation}
that is then connected to an LSTM \cite{Hochreiter1997LONGMEMORY} layer that can store and use previous features
\begin{equation}
\bm{h}^{(4)}_t = \text{LSTM}\left( \bm{h}^{(3)} | \bm{h}^{(4)}_{t-1} \right).
\end{equation}
The output from the LSTM is then sent through a final layer
\begin{equation}
\bm{Q} = (\bm{W}_Q \bm{h}^{(4)}+ \bm{b}_4)\circ\bm{Q}_\mathrm{mask},
\end{equation}
where the operator $\circ$ denotes pointwise multiplication, $\bm{Q}_\mathrm{mask}$ is a masking vector described in the next section, $\bm{W}_Q$ and $\bm{b}_4$ are the weights and biases for the final layers, respectively. The optimal policy $\pi^\star$ is then given by maximizing the optimal action value function $\bm{Q}^\star$ as
\begin{equation}
\pi^\star(s_t) = \arg\max_{a_t} \bm{Q}^\star(s_t,a_t).
\end{equation}

%

\subsection{Q-masking}
\label{sec:masking}
\noindent
Q-masking \cite{Mukadam2017} helps the learning process by reducing the exploration space by disabling actions the agent does not need to explore. If there are less than $N$ cars, it would then be meaningless to choose to follow a car that does not exist. Which motivates masking off cars that do not exist. In previous work \cite{Tram2018LearningQ-Learning}, a high negative reward was given when an action to follow a car that did not exist was chosen, while the algorithm continued with a default action take way. The agent quickly learned to not choose cars that did not exist, but with Q-masking, the agent does not have to explore these options. For further details about the training see \cite{Tram2018LearningQ-Learning}.

\subsection{Simulation environment}
\noindent
All agents are spawned with a random intention, initial speed $v_0 \in [10,30]$m/s, and position $p^i_0 \in [10,55]$m. The vehicle dimensions are $2$ m wide and $4$ m long. The ego car operates within comfort bounds and therefore has a limited maximum acceleration and deceleration of $5$ m/s$^2$. 
\begin{figure}[t]
	\centering
	\vspace{0.2cm}
	\includegraphics[width=.8\columnwidth]{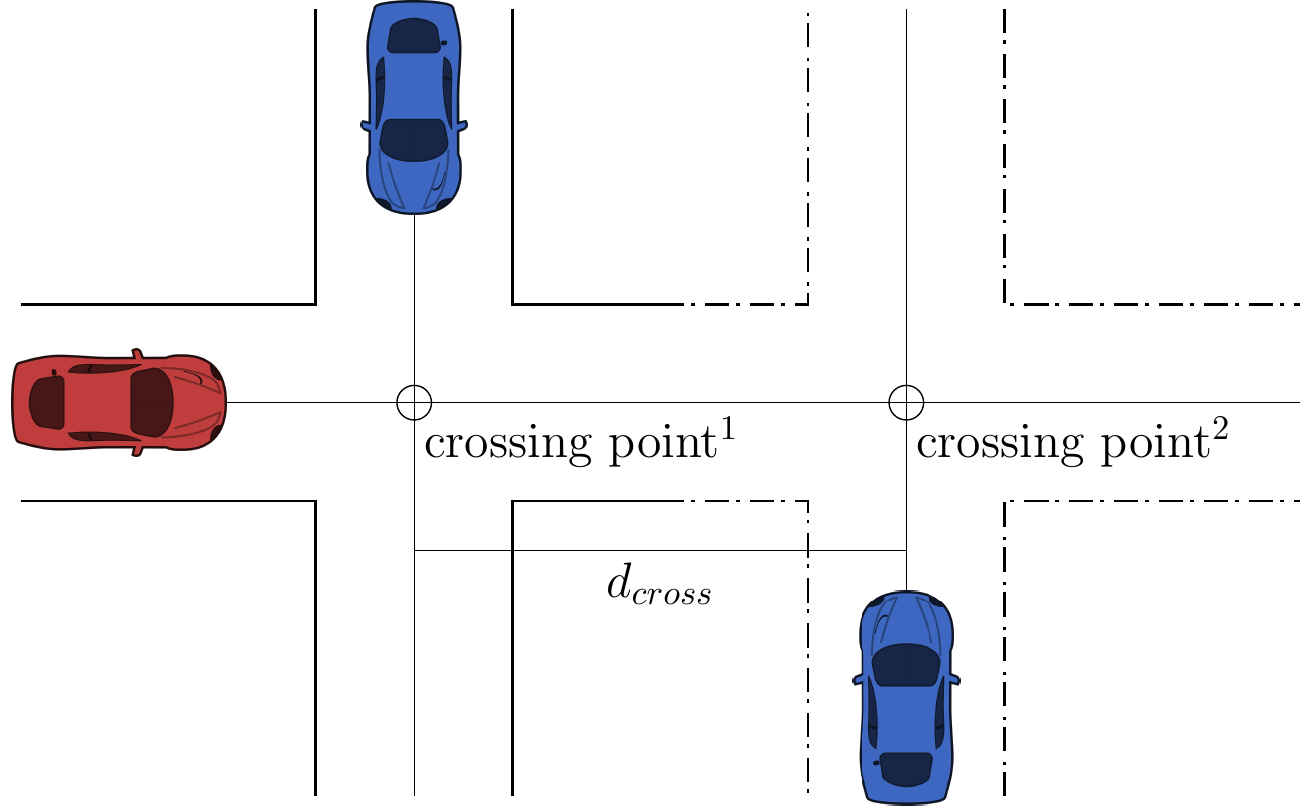}
	\caption{Illustration of a intersection scenario, where the solid line is a single crossing and together with the dashed line creates a double crossing.}
	\label{fig:scenario}
	\vspace{-0.2cm}
\end{figure}
Two main types of crossing were investigated. One and two crossing points as shown in Fig. \ref{fig:scenario}, where the distance $d_{cross}$ between crossing points vary between $[4, 8, 12, 25, 30, 40]$m with each scenarios. 

The MPC agent was discretized at $30$Hz, with a prediction horizon of $N=100$ and cost tuning of
\begin{gather}
\bar{Q}=\mathrm{blockdiag}(0.0,1.0,1.0),\ \bar{R}=1,\ \bar{S}=\mathbf{0}.
\end{gather}

\subsection{Reward function tuning}\label{sec:reward}
\noindent
There are three states that terminates an episode: success, failure, and timeout. Success is when the ego agent reaches the end of the road defined by the scenario. Failure is when the frame of the ego agent overlaps with another road users' frame, e.g., in a collision, this frame can be the size of the vehicle or a safety boundary around a vehicle. The final terminating state is timeout, which is simply when the agent can not reach the two previous terminating states before the elapsed time $\tau$ reaches the timeout time $ \tau_m$.

According to \cite{VanHasseltLearningMagnitude}, the $Q_\pi$ values and gradient can grow to be very large if the total reward values are too large. All rewards are therefore scaled with the episode timeout time $\tau_m$, which is set to 25$s$, to keep the total reward $r_t \in [-2, 1]$.
The reward function is defined as follows:
\begin{align*}
r_t = &\begin{cases}
1 & \text{on success, }\\
-1                & \text{on failure},\\
0.5                 & \text{on timeout, i.e. } \tau \ge \tau_m,\\
f(p_{\text{crash}}, p_{\text{comf}}) & \text{on non-terminating updates},\\
\end{cases} 
\end{align*}
where $f(p_{\text{crash}}, p_{\text{comf}})$ consists of
\begin{equation}
f(p_{\text{crash}}, p_{\text{comf}})  = \alpha p_{\text{crash}}  \frac{\tau_m}{\tau - t_{pred}} + \beta p_{\text{comf}} \frac{\tau_m}{\tau},
\label{eq:mpc_reward}
\end{equation}

\noindent
with  $\alpha\in[0,1]$, $\beta\in[0,1]$ being weight parameters, and $\alpha+\beta=1$. The first term $p_\mathrm{crash}$ corresponds to a feasibility check of Problem \eqref{eq:mpc}, which to a large extent depends on the validity of the accuracy of the prediction layer. The high-level decision from the policy-maker affects how the constraints are constructed, and may turn the control problem infeasible, e.g. if the decided action is to take way, while not being able to pass the intersection before all other obstacles. Therefore, whenever the MPC problem becomes infeasible we set $p_\mathrm{crash}=1$, otherwise $p_\mathrm{crash}=0$, to indicate that the selected action most likely will result in a collision with the surrounding environment. Because $p_{\text{crash}}$ usually only triggers close to a potential collision, $t_{pred}$ is set to the first time a crash prediction is triggered, to scale the negative reward relatively higher the later it is predicted. 

The second term $p_\mathrm{comf}$ relates to the comfort of the planned trajectory, which is estimated by computing and weighting the acceleration and jerk profiles as
\begin{align*}
p_\mathrm{comf} = \frac{1}{\sigma{}N}(\sum_{k=0}^{N-1} \bar{a}_k^2\bar{Q}^a + \bar{j}_k^2\bar{R}^j + \bar{a}_N^2\bar{Q}^a),
\end{align*}
where $\bar{a}$, and $\bar{j}$ are the acceleration component of the state and jerk component of the control, respectively, $Q^a$ and $R^j$ are the corresponding weights, and $\sigma$ is a normalizing factor which ensures that $p_\mathrm{comf}\in[0,1]$. For the simulation we used $\bar{Q}^a=1$ and  $\bar{R}^j=1$.

The timeout reward 0.5 was set to be higher than the average accumulated reward from $p_\mathrm{comf}$, so that the total accumulated reward would be positive in case of timeouts. 

\section{Results}\label{sec:results}
\noindent
For evaluation we compared the success rate of the decision-policy together with a collision to timeout ratio~(CTR). The success rate is defined as the number of times the agent is able to cross the intersections without colliding with other obstacles or exceeding the time limit to cross. Since we define a time-out to be a failure, we use the CTR to separate potential collisions from the agent being too conservative.

\begin{figure}[t!]
	\centering
	\includegraphics[width=0.9\columnwidth]{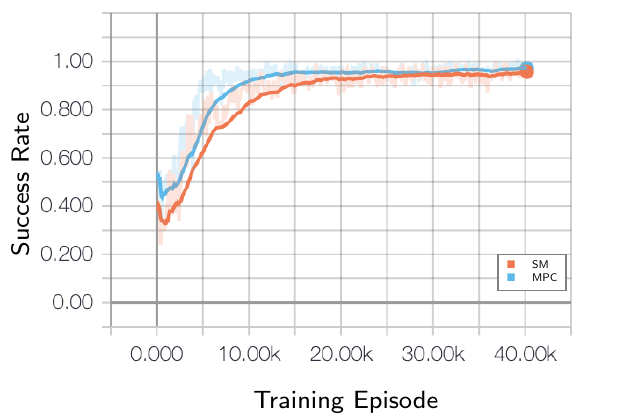}
	\caption{Average MPC and SM success rate for a single corssing after evaluating the policy 300 episodes.}
	\label{fig:result1}
\end{figure}

Fig. \ref{fig:result1} shows a comparison in success rate between the proposed MPC architecture and the previous SM agent for scenarios with only one intersection. In this scenario, the MPC agent converges after $10^4$ training episodes, while the previous SM agent converges after $4\cdot{}10^4$ training episodes. In addition, comparing the CTR metric, Tabel. \ref{tab:successrate} shows that the MPC agent has CTR of $0.45$, while the SM agent has a CTR of $0.72$.
Evidently, it is apparent that the MPC is able to leverage future information into its planning horizon in order to achieve faster training, higher success rates, and also avoiding more collisions as a result.

We evaluate the performance of the MPC and SM agents for the more difficult double intersection problem, where we vary the distance between the intersection points. Table~\ref{tab:successrate} shows the performance of the MPC and SM agent for both the single and double scenarios. The performance decreases for both agents in the double crossing scenario. However, it is evident that the MPC agent suffers less performance degradation compared to the SM agent. The CTR more than doubles for the MPC agent for the double crossing, while the already high CTR rate for the SM agent increases above rates of $0.9$. Still, the MPC agent manages  to outperform the SM agent.

\begin{table}[t!]
	\centering
	\caption{Average success rates and collision to timeout rates.}
	\begin{tabular}{ p{1,3cm}|p{0,8cm}|p{0,8cm}|p{0,8cm}|p{0,8cm}}
	Controller &\multicolumn{2}{c|}{Success Rate}
	&\multicolumn{2}{c}{CTR}\\
	\hline
	 & Single & Double & Single & Double\\
	SM & $96.1\%$ & $90.9\%$ & $72\%$ & $93\%$\\
	MPC & $97.3\%$ & $95.2\%$ & $45\%$ & $76\%$\\
\end{tabular}
\vspace{-2em}
\label{tab:successrate}
\end{table}
\section{Discussion}\label{sec:discussion}
\noindent
The benefit of being able to use a prediction horizon for the MPC is shown to mostly impact the training time for the traffic scenarios compared to the SM agent. This allows the RL decision-policy to get feedback early in the training process to see whether an action most likely will lead to a collisions. In addition, the lower CTR also implies that the use of a prediction horizon also makes the decision-policy more conservative, since it rather times out than risk collisions.

It is important to note that only little effort was put into tuning the MPC agent, and that we used very primitive prediction methods that do not hold very well in crossing scenarios, e.g. the simulated agents did not keep constant speed profiles while approaching the intersections. However, under these circumstances, the decision algorithm still managed  to obtain a success rate above $95\%$ for the double crossings.

In practice, a full decision architecture system would include a safety layer that limits which acceleration values the system can actuate in order to stay safe, followed by the decisions algorithm from this work that generates an acceleration request. The environment state, together with the new acceleration request, could be sent through a collision avoidance system that checks if the current path has a collision risk, and avoiding collision by allowing higher acceleration limits. This way, a failure would correspond to a intervention by the collision avoidance system instead of a crash.

\section{Conclusion}\label{sec:conclusion}
\noindent
In this paper, we proposed a decision making algorithm for intersections which consists of two components: a high-level decision maker that uses Deep Q-learning to generate decisions for how the vehicle should drive through the intersection, and a low-level planner that uses MPC to optimize safe trajectories. We tested the framework in a traffic simulation with randomized intent of other road users for both single and double crossings. Results showed that the proposed MPC agent outperforms the previous SM agent with $95.2\%$ success rate in scenarios with double crossings compared to $90.9\%$ for the SM agent. Results also showed that the crash timeout ratio was also significantly lower at $45\%$ for the MPC agent compared to $72\%$ for the SM agent in single crossings. Meaning, the proposed method is better at handling scenarios with multiple intersections and vehicles. 

\bibliographystyle{IEEEtran} 
\bibliography{mendeley}

\end{document}